\documentclass[11pt, letterpaper]{article}

\usepackage{amsmath}
\usepackage{amssymb}
\usepackage{amsthm}
\usepackage{graphicx}
\usepackage{url}
\usepackage{algorithm}
\usepackage{algorithmicx}
\usepackage{algpseudocode}
\usepackage{caption}
\usepackage{subcaption}
\usepackage{booktabs}
\usepackage{authblk}
\usepackage{appendix}
\usepackage[utf8]{inputenc}
\usepackage[T1]{fontenc}
\usepackage[english]{babel}
\usepackage{geometry}
\usepackage{xcolor}
\usepackage{fancyhdr}
\usepackage{microtype}
\usepackage{parskip}
\usepackage{tabularx}
\usepackage{enumitem}
\usepackage{cite}
\usepackage{longtable}
\usepackage{hyperref}

\hypersetup{
    colorlinks=true,
    linkcolor=blue,
    filecolor=magenta,      
    urlcolor=cyan,
    citecolor=red,
}

\definecolor{headercolor}{RGB}{0, 50, 100}
\title{PolicyGPT: Automated Analysis of Privacy Policies with Large Language Models}

\author[1]{Chenhao Tang}
\author[2]{Zhengliang Liu}
\author[3]{Chong Ma}
\author[2]{Zihao Wu}
\author[2]{Yiwei Li}
\author[4]{Wei Liu}
\author[5]{Dajiang Zhu}
\author[6]{Quanzheng Li}
\author[6]{Xiang Li}
\author[2]{Tianming Liu}
\author[1]{Lei Fan}

\affil[1]{School of Cyber Science and Engineering, Shanghai Jiao Tong University, Shanghai 200240, China}
\affil[2]{School of Computing, The University of Georgia, Athens 30602, USA}
\affil[3]{School of Automation, Northwestern Polytechnical University, Xi'an 710072, China}
\affil[4]{Department of Radiation Oncology, Mayo Clinic, Phoenix 85054, USA}
\affil[5]{Department of Computer Science and Engineering, The University of Texas at Arlington, Arlington 76019, USA}
\affil[6]{Department of Radiology, Massachusetts General Hospital and Harvard Medical School, Boston 02114, USA}

\begin{document}

\maketitle

\begin{abstract}

Privacy policies serve as the primary conduit through which online service providers inform users about their data collection and usage procedures. However, in a bid to be comprehensive and mitigate legal risks, these policy documents are often quite verbose. In practical use, users tend to click the \textbf{Agree} button directly rather than reading them carefully. This practice exposes users to risks of privacy leakage and legal issues. Recently, the advent of Large Language Models (LLM) such as ChatGPT and GPT-4 has opened new possibilities for text analysis, especially for lengthy documents like privacy policies. In this study, we investigate a privacy policy text analysis framework \textbf{PolicyGPT} based on the LLM. This framework was tested using two datasets. The first dataset comprises of privacy policies from 115 websites, which were meticulously annotated by legal experts, categorizing each segment into one of 10 classes. The second dataset consists of privacy policies from 304 popular mobile applications, with each sentence manually annotated and classified into one of another 10 categories. Under zero-shot learning conditions, \textbf{PolicyGPT} demonstrated robust performance. For the first dataset, it achieved an accuracy rate of 97\%, while for the second dataset, it attained an 87\% accuracy rate, surpassing that of the baseline machine learning and neural network models.

\end{abstract}

\section{Introduction}

The trend of expanding privacy policies has been gaining momentum, with a significant surge especially noticeable subsequent to the unveiling of the General Data Protection Regulation (GDPR) by the European Union. This comprehensive piece of legislation, designed to upgrade privacy standards and grant individuals unprecedented control over their personal data, has triggered widespread changes across the digital landscape. In the wake of GDPR's introduction, a substantial proportion of websites, approximately 72.6\%, have taken the initiative to revise and update their privacy policies\cite{degeling2018we}. This statistic underscores the sweeping impact of the regulation on the digital sphere and the consequent efforts by website operators to ensure compliance with its provisions. A year following the enforcement of GDPR, an interesting pattern has emerged in terms of privacy policy lengths. Specifically, websites operating within the bounds of the EU have witnessed an increase of 35.39\% in the textual length of their privacy policies. This substantial increase in length likely reflects the inclusion of more detailed and comprehensive disclosures, as mandated by GDPR, to ensure that users are fully informed about how their data will be used. Simultaneously, this trend of elongating privacy policies is not confined to the European Union. In fact, on a global scale, privacy policies have undergone a sizeable expansion as well, with their textual length experiencing an increase of 25.21\%\cite{linden2018privacy}.

Examining the issue from the viewpoint of users, the situation significantly exacerbates the already time-consuming process of perusing privacy policies\cite{mcdonald2008cost, amos2021privacy}. These policies, often steeped in legal jargon and dense text, require considerable time and understanding to fully comprehend. The complexity and length of these documents can be daunting, leading to an extended reading process that can be both tedious and overwhelming. In light of this daunting task, a growing number of users are finding themselves inclined towards an easier route: directly clicking the \textbf{Agree} button. This action, often executed in haste and without sufficient consideration, bypasses the need to understand the intricate details embedded within these policies. Users, in their eagerness to proceed, may not fully consider the nature and extent of the information that is being collected from their actions by the website or application. This includes, but is not limited to, browsing habits, personal preferences, location data, and other forms of digital footprints that can be tracked and stored. Furthermore, the understanding of how and where to revoke consent or disable certain options is often neglected. These controls, which are integral to managing personal data and maintaining digital privacy, are often buried deep within settings or masked by confusing terminologies. As such, users may not know how to navigate these options or even be aware that they exist. This lack of understanding further compounds the privacy issues associated with the hasty acceptance of these policies.

The recent emergence of large language models (LLMs), exemplified by ChatGPT and GPT-4, presents new opportunities for potential advancements in this field. The LLMs exhibit impressive capabilities in text analysis, thanks to their ability to understand and generate human-like text based on the patterns they learned during their training phase on a massive corpus of data. These advancements in language understanding and text generation could significantly augment the process of privacy policy analysis, leading to more accurate and efficient categorization. Despite the promising potential of LLMs, methods based on these models are still in their infancy as of now. To the best of our knowledge, this research study represents the pioneering effort to investigate the application of LLMs specifically for privacy policy analysis and categorization. In this study, we introduce a new model, referred to as \textbf{PolicyGPT}, which leverages the capabilities of ChatGPT. The operation of this model is bifurcated into two primary steps. Initially, we formulate the task content and establish the definitions of various categories. This step is crucial as it lays the groundwork for the classification task. Subsequently, we proceed to the second step where we supply the text that requires categorization, along with a prompt, to ChatGPT or GPT-4. The prompt provides necessary context for the model to understand its task. Under the guidance of the established task content and category definitions, ChatGPT or GPT-4 then processes the input text. Based on its understanding and the provided context, the LLM assigns the most appropriate category to the input text. Upon obtaining the results, we juxtaposed them with the classifications annotated manually and computed the accuracy. The data indicates that our framework, even zero-shot, has demonstrated superior performance compared to that of existing works.

\section{Related Work}

\subsection{Privacy Policy Analysis}

With the proliferation and enhancement of privacy policies for webpages and applications, research on privacy policies is experiencing an explosive growth. As early as 2008, McDonald et al. conducted a study involving 212 participants reading privacy policies of different lengths and estimated through modeling that if US Internet users read online privacy policies word for word, they would need to spend 201 hours annually, with a time cost exceeding \$700 billion\cite{mcdonald2008cost}. This was in an era 15 years ago when an individual only visited an average of 119 webpages per year.

After the introduction of the General Data Protection Regulation (GDPR) by the European Union, many related studies have been initiated focusing on comparing the privacy policies before and after the GDPR. Linden et al. created a diverse corpus that contains 6278 unique English privacy policies from both within and outside the EU, including versions before and after GDPR \cite{linden2018privacy}. Their analysis revealed that both EU and global privacy policies have noticeably lengthened, and the coverage of topics highly relevant to GDPR in the policies has significantly improved. The privacy policies have become more specific in describing their data practices. In the study conducted by Degeling et al., they scanned 500 popular websites from each of the 29 European Union member states\cite{degeling2018we}. Their findings indicated that 15.7\% of these websites incorporated new privacy policies, and 84.5\% of the websites had existing privacy policies. Among the websites with existing privacy policies, 72.6\% updated their policies in proximity to the effective date of the General Data Protection Regulation (GDPR). They concluded that upon the implementation of GDPR, the internet environment has experienced an increase in transparency. However, it still lacks effective and user-friendly mechanisms to enable users to either consent to or refuse the processing of their personal data on the internet.

In the realm of identifying, extracting, and analyzing privacy policies, numerous works have introduced a variety of datasets to gauge extraction and analysis efficacy. In 2014, Ramanath et al. proposed a dataset that consisted of over 1000 manually annotated entries\cite{ramanath2014unsupervised}. Wilson et al. introduced OPP-115, a dataset composed of 115 website privacy policies, manually annotated by several legal professionals, and containing over 3000 segment-level annotations in 2016\cite{wilson2016creation}. Zimmeck et al. proposed APP-350, a dataset that includes 350 annotated mobile application privacy policies\cite{zimmeck2019maps} in 2019. In 2020, Bannihatti introduced the Opt-out Choice Dataset, featuring over 1000 sentence-level entries with human annotated tags\cite{bannihatti2020finding}. Subsequently, in 2021, Nokhbeh and his team utilized DMOZ (a vast open content directory on the internet) and its manually classified 1.5 million websites to gather hundreds of thousands of privacy policies related to their categories\cite{nokhbeh2021large}, thereby enabling the study of privacy policies across different categories or market sectors. In the same year, Amos and his colleagues, employing a web crawler and adhering to a series of validation and quality control steps, compiled a dataset comprising 1,071,488 English privacy policies \cite{amos2021privacy}. This dataset covers a span of over twenty years and encompasses more than 130,000 distinct websites. Also in the same year, Bui et al. created a large dataset containing 4.1k sentences (97k tokens) and 2.6k fine-grained annotated data practices from 30 real-world privacy policies, aimed at training and evaluating neural networks \cite{bui2021automated}. Concurrently, Liu et al. collated a corpus comprising 36,610 tagged sentences from privacy policies of 304 mobile device applications \cite{liu2021have}. These privacy policies were divided into sentences, which were manually categorized into ten classes. Every sentence was independently annotated by three volunteers. If the three annotations were identical, that annotation was considered final for the sentence. If the annotations differed, they would engage in discussions until consensus was reached. In 2022, Rahman et al. utilized a Python-based scraping tool to extract data from the Google Play Store. They collected meta-information and privacy policies from 213,000 application. They then extracted the AndroidManifest.xml files, which declare permissions, from the APKs and established a dataset of application permissions\cite{rahman2022permpress}.

Studies employing various models and methods have been conducted based on these datasets. In 2018, Harkous et al. proposed an automated framework for privacy policy analysis, Polisis, built upon 130K privacy policies and a novel hierarchical structure of neural network classifiers, achieving an accuracy of 88.4\% on the OPP-115 dataset\cite{harkous2018polisis}. In addition, they developed a free-form question-answering system for privacy policies, PriBot, which provided correct answers within the top three results for 82\% of the test questions. Zimmeck et al. used Support Vector Classification (SVC), a mechanism within Support Vector Machines (SVM), on their self-generated APP-350 dataset, reaching an average F1 score of 0.71\cite{zimmeck2019maps}. Bannihatti et al. used Logistic Regression and BERT for classification on their collected OPT-out dataset, with F1 scores ranging from 0.5 to 0.85 and 0.6 to 0.9 respectively\cite{bannihatti2020finding}. However, they claimed that the classification performance could be enhanced to exceed 0.9 by incorporating some readily identifiable OPT-out instances. Sathyendra et al. also used the OPP-115 dataset in 2017 \cite{sathyendra2017identifying}. They proposed a two-phase classification model architecture for identifying OPT-out options in privacy policy text, achieving an average F1 score of 0.735. Liu et al. utilized their own collected and annotated PPGDPR dataset\cite{liu2021have}. They employed three models — SVM, BiLSTM, and BERT — to measure the performance of sentence classification tasks, achieving average F1 scores of 0.505, 0.643, and 0.717 respectively.

\subsection{Large Language Model}

Large language models have recently emerged as a powerful approach for natural language processing \cite{liu2023summary,zhao2023brain,zhao2023survey}. Transformer-based \cite{vaswani2017attention} language models are pretrained on massive amounts of text data, with hundreds of billions or more parameters. Notable LLMs include models such as GPT-3 \cite{brown2020language}, PaLM \cite{chowdhery2022palm}, and GPT-4 \cite{openai2023gpt}.

A defining characteristic of LLMs is that they exhibit surprising abilities not present in smaller models, often referred to as emergent abilities \cite{wei2022emergent}. For instance, GPT-3 demonstrated strong few-shot learning through in-context examples \cite{brown2020language}, while PaLM \cite{chowdhery2022palm} showed improved generalization when tuned on diverse instructions. It is speculated that LLMs acquire such abilities once model scale exceeds a sufficient level \cite{wei2022emergent,liu2023summary}.

LLMs performance typically improves with increased model size, data size, and compute \cite{kaplan2020scaling,hoffmann2022training}. Key techniques for developing LLMs include scaling, optimized distributed training, prompting strategies to elicit abilities, and alignment tuning to improve safety \cite{zhao2023survey}. Applications of LLMs span domains like natural language processing, information retrieval, computer vision \cite{gupta2023visual}, and healthcare \cite{li2023chatgpt,liu2023deid,ma2023impressiongpt}.

\section{Datasets}

In this chapter, we elucidate the datasets, OPP-115\cite{wilson2016creation} and PPGDPR\cite{liu2021have}, deployed in our study and delineate the preprocessing techniques applied to them. The datasets we have selected for our study are related to privacy policies of two distinct digital platforms: web and mobile. Each dataset has unique characteristics that make it suitable for our analysis.

The first dataset is focused on web-based privacy policies. These policies are fundamental to the operation of various online platforms and services. They dictate how user data is collected, stored, and shared, making them a key area of interest for our research. The second dataset pertains to privacy policies designed for mobile platforms. With the ever-increasing use of mobile applications, understanding the nuances of mobile privacy policies has become increasingly relevant. These policies often differ from web-based ones due to the unique nature of mobile data collection and usage. Additionally, the ten labels for this dataset were derived from Article 13 of the GDPR. Several years after the implementation of the GDPR, studies investigating the connection between privacy policies and the GDPR are of significant value.

A critical aspect of both these datasets is the high degree of reliability they offer. This is attributable to the fact that they have been manually annotated by professional legal and computer experts. Manual annotation, especially by trained professionals, provides a level of accuracy and detail that automated processes might fail to achieve. This meticulous process of annotation therefore bolsters the validity of our analysis and findings.

Additionally, these datasets have been the subject of existing research, providing a solid foundation of knowledge and context for our study. This pre-existing body of research allows us to better interpret our results and draw more informed conclusions. Therefore, given their high reliability and the comprehensive insights they offer, we have opted to utilize these two datasets for our research study.
Our work primarily focuses on the discussion and comparison of privacy policy categorization at the segment level.


\subsection{Categories}

The OPP-115 dataset is a comprehensive compilation of privacy policies, sourced from a diverse selection of 115 different websites. This web-based focus is significant, as it provides a snapshot of privacy practices across the digital landscape, capturing the breadth of data management approaches utilized by online entities. Each privacy policy included in the dataset has been meticulously annotated and categorized by a team of legal experts. Given the intricate and often complex nature of web-based privacy policies, this expert involvement is crucial. It ensures that the dataset's annotations accurately reflect the nuanced contents of these policies, and that the categorization is grounded in a solid understanding of legal and data privacy principles. This contributes to the accuracy and reliability of the dataset, making it a dependable resource for research into online privacy matters. The categorization system employed within OPP-115 is particularly extensive. It breaks down the data into ten distinct classes. This level of granularity is pivotal when examining web-based privacy policies, as it allows for a detailed analysis of various aspects of data handling practices. The specific categories and their descriptions among the datasets are showed in Table \ref{table-categories-opp-115}.

Conversely, PPGDPR, our second dataset, is an aggregation of privacy policies gathered from a selection of 304 apps on the Google Play Store. Similar to OPP-115, a team of legal and computer science experts has carefully annotated these privacy policies, marking up each policy to identify and categorize various privacy-related aspects and concerns. The classification structure in the PPGDPR dataset derives its foundation from Article 13 of the General Data Protection Regulation (GDPR), which lends the dataset a high degree of standardization and formality. This structuring is not arbitrary; it is deeply rooted in the legal framework of GDPR, ensuring the precision and relevance of the classification criteria. The primary focus of this dataset's classification is the rights of the users. This is a significant aspect as it underscores the importance of user privacy and data protection, reflecting the spirit of GDPR. The user rights-centric approach of this dataset aligns with the modern emphasis on personal data sovereignty, and it signifies the commitment to respect and uphold the autonomy of individuals in the context of data usage and protection. Contrasting with OPP-115, this dataset conducts annotation and categorization on a sentence-by-sentence basis, with each sentence having only one annotation. This is due to the fact that when the individuals annotating a sentence encounter disagreement, they engage in discussion until a consensus is reached. The specific categories and their descriptions among the datasets are showed in Table \ref{table-categories-ppgdpr}.

\subsection{Preprocess}

\textbf{Policy Extraction} The process of extracting privacy policy text involves several crucial steps, which are primarily centered around the use of web crawling technologies and strategic content filtering. Usually, these texts are embedded within well-structured and aesthetically pleasing webpages, adding an element of complexity to the extraction process.

A common tool utilized for webpage crawling and data storage is the Scrapy Web framework. This open-source and collaborative framework offers a comprehensive toolkit for extracting the data needed from websites, rendering it particularly useful in the context of privacy policy extraction. It facilitates the process of navigating through the website, identifying the necessary information, and saving it for further processing. However, due to the intricate designs of these webpages and the heavy reliance on Javascript for loading content, the extractor must patiently wait for the entire page to be fully loaded and for all Javascript scripts to execute completely before initiating the extraction process. This step is crucial to ensure no pertinent information is missed during the extraction process.

The next stage involves the removal of all unnecessary elements from the webpage that are unrelated to the privacy policy. This includes the HTML headers, footers, menu pages, and styles. The aim of this step is to distill the webpage content down to only the essential components, thus eliminating any potential noise or irrelevant information that may detract from the core privacy policy text. Following the filtering process, the resulting HTML consists solely of the privacy text body. This text is typically marked with <p> tags, representing individual paragraphs of the policy. Additionally, newline symbols, represented by <br> tags, are retained to maintain the original formatting and readability of the text.

\textbf{Policy Segmentation} The process of segmenting privacy policy texts can be a crucial aspect of subsequent analysis or processing. This segmentation can be performed in two primary ways: sentence-wise and paragraph-wise, each presenting its own nuances and challenges.

Segmenting the text into sentences is a relatively straightforward task. It essentially involves dividing the text at every instance of a full stop or period. This form of segmentation is simple, as it mostly relies on a consistent grammatical rule: sentences typically end with a full stop. This process results in a list of all the sentences present in the privacy policy, effectively breaking down the text into its most basic coherent units.

\begin{figure}[!ht]
  \centering
  \includegraphics[width=0.8\textwidth]{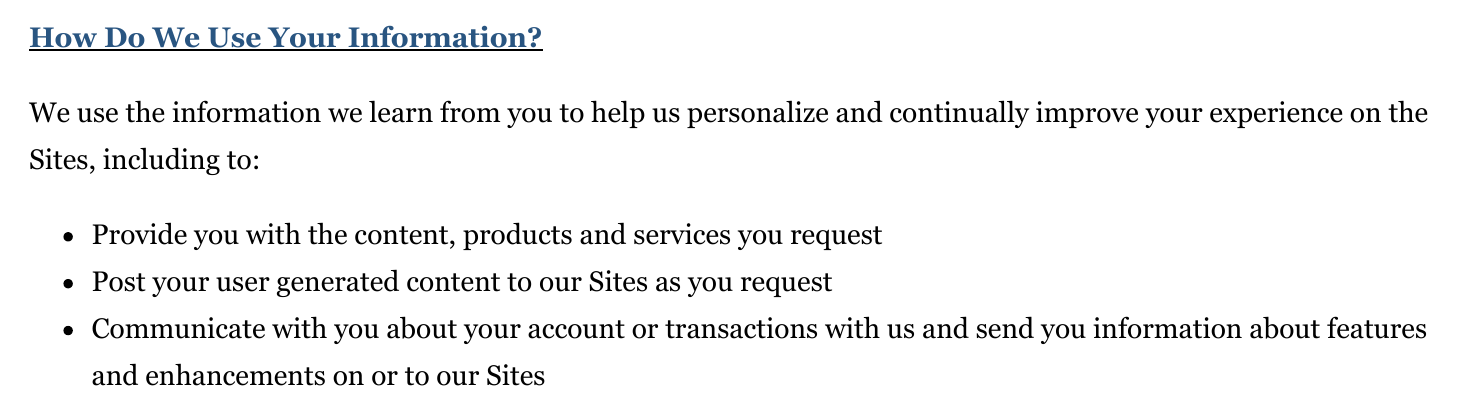}
  \caption{Typical Privacy Policy Structure.}
  \label{opp-115-preprocess}
\end{figure}

On the contrary, paragraph-wise segmentation of privacy policy texts is more complex. Within such texts, there are often several instances where a block of text enumerates an ordered or unordered list, as illustrated in Figure \ref{opp-115-preprocess}. In these situations, it would be incorrect to treat each list item as a standalone paragraph. This is because the elements in the list are usually interconnected and often rely on the explanatory text above the list to be fully understood.For instance, a privacy policy might list various types of information the company collects, followed by a list of ways in which this information is used. Each list item does not provide a complete idea on its own and must be read in conjunction with the introductory text to fully understand the context.

Therefore, when segmenting the text into paragraphs, it is vital to consider the context and structure of the information. In general, list items should be merged with the preceding text into a single paragraph. This approach ensures that all the information related to a particular topic is grouped together, facilitating a more accurate understanding and analysis of the privacy policy.




\section{Method}

\subsection{Prompt Generation}

\begin{figure}[!h]
  \centering
  \includegraphics[width=0.8\textwidth]{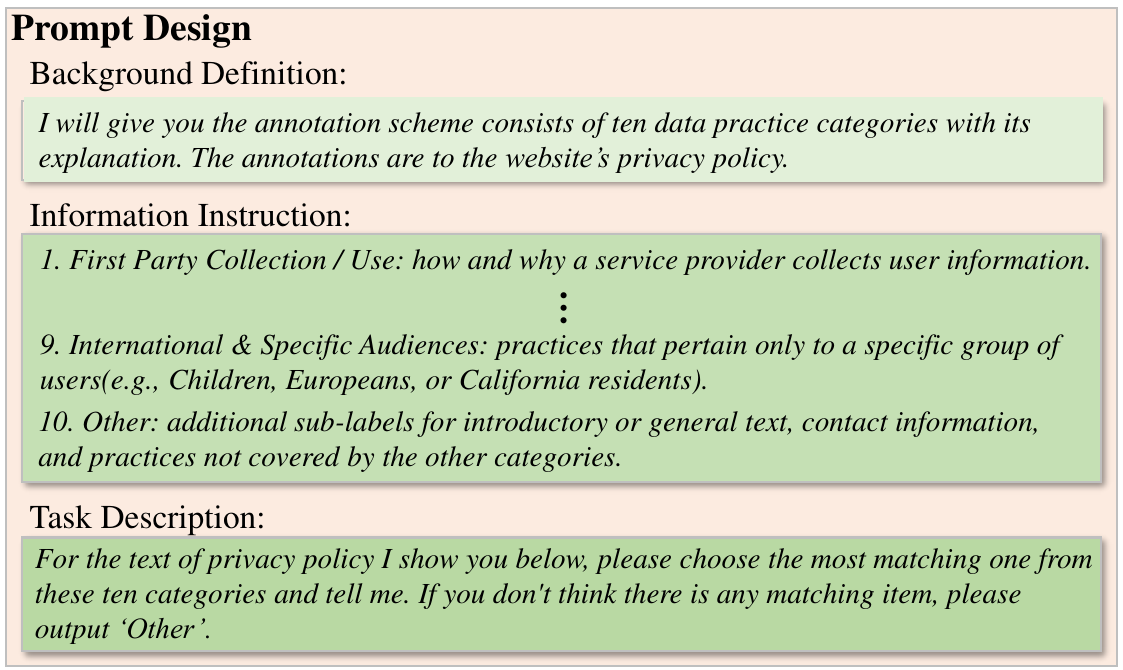}
  \caption{Details of our prompt design.}
  \label{prompt_design}
\end{figure}

In the context of this study, we implemented a design strategy known as the "prefix prompt" to enhance the model's capability in understanding the complexities of privacy classification and definition. Our prompt is structured into three distinct segments, which include the background definition, information instruction, and task description.

As exemplified in Figure~\ref{prompt_design}, we used the OPP-115 dataset as a model to illustrate the practicality of our approach. The task description component begins by providing a contextual backdrop for the task at hand, which is delineated based on ten categories related to privacy. Following this, we introduce the names and individual descriptions of each category, a step that empowers the model to learn and grasp the nuanced meanings embedded in each privacy category. The task description component is constructed by integrating a question with the target text. In the question section, the task is introduced. This task requires the model to engage in a thorough analysis of the target text and subsequently generate a classification result. This result is determined by referencing the category descriptions provided in the information instruction component of the prompt. The target text is then attached to the end of the prompt, completing the structure.

In relation to the PPGDPR dataset, we adopted an identical prompt design as demonstrated in Figure~\ref{prompt_design}. This entailed the incorporation of an additional ten privacy classifications along with their corresponding descriptions as provided by the dataset itself. This approach ensures that the model is well equipped to handle a diverse range of privacy classifications, thereby demonstrating the versatility and adaptability of our prefix prompt design.

In our view, the use of a prefix prompt is pivotal in enabling LLM to effectively absorb the semantic information pertaining to privacy categories and the target text. This, in turn, boosts its proficiency in undertaking classification tasks. The complete prompts can be referred to in Section \ref{sec:full-prompt}.


\subsection{Baseline Models}

\textbf{ChatGPT}
ChatGPT (Generative Pre-training Transformer) is a conversational generation model developed by OpenAI, based on natural language processing techniques and neural network models. It is trained using a large amount of text data and self-supervised learning techniques, enabling it to analyze and reason based on specific contexts and generate high-quality conversations. ChatGPT can also perform text classification tasks and generate responses that align with given questions based on contextual semantics. It has been extensively integrated into diverse applications, including education and healthcare, and exhibits strong performance in tasks such as text classification, data augmentation, summarization, and other natural language processing tasks.

\textbf{GPT-4}
GPT-4, the latest architecture released by OpenAI, is capable of handling both text and image modalities, in addition to processing regular text information like ChatGPT. It exhibits enhanced reliability and a greater understanding of subtle semantic instructions compared to ChatGPT when handling more complex tasks. Based on the GPT-3.5 architecture, GPT-4 incorporates RLHF (Reinforcement Learning from Human Feedback) techniques to further train the model through human feedback on its outputs. GPT-4 outperforms ChatGPT across various tasks, making it an important baseline model for comparison in our study.

\textbf{Claude2}
Claude2 is a large-scale language model developed by Anthropic, the latest version to their model. Similar to the ChatGPT/GPT-4 models, Claude2 utilizes the transformer architecture and has been trained using unsupervised learning and RLHF techniques. One notable feature of Claude2 is its ability to support text inputs of up to 100,000 tokens, surpassing the 32,000 tokens supported by GPT-4. This indicates its enhanced capacity for context analysis and processing. Therefore, Claude2 serves as an important baseline model for our comparative analysis.

\textbf{PaLM}
PaLM is a large language model developed by Google, based on the Pathway training architecture. Unlike other large language models, the Pathway architecture integrates multiple independent tasks, enabling it to comprehend various forms of data input and achieve efficient training simultaneously. It can efficiently train on thousands or tens of thousands of acceleration units. Furthermore, PaLM has achieved state-of-the-art few-shot results in hundreds of natural language, code, and mathematical reasoning tasks. Therefore, we have chosen PaLM as our baseline model.

\textbf{LLaMA2}
LlaMA2 is a large language model developed by Meta. It is an optimized auto-regressive language model trained using supervised fine-tuning and RLHF techniques. Specifically, LlaMA2-Chat is a variant of LlaMA2 that is tailored for conversational scenarios. It outperforms open-source large language models in most benchmarks and can even surpass closed-source models like GPT-4 on certain test sets. The availability of an open-source license for commercial applications has also made LlaMA2 a recent hot-spot of interest. Therefore, we have also introduced LlaMA2 as a baseline model. 

\subsection{Implementation}

The ChatGPT model used in this paper is 'gpt-3.5-turbo-0613', released on June 13th, 2023, and the GPT-4 model employed in this paper is 'gpt-4-0314', released on March 14th, 2023. The Anthropi Claude model evaluated in this study is 'claude-2'. 

\subsection{Evaluation Metrics}

True Positives (TP) are the cases where the model correctly predicted the positive class.
True Negatives (TN) are the cases where the model correctly predicted the negative class.
False Positives (FP) are the cases where the model incorrectly predicted the positive class.
False Negatives (FN) are the cases where the model incorrectly predicted the negative class.

\textbf{Accuracy} is a metric used in machine learning that measures the overall correctness of a classification model. It is defined as the ratio of the number of correct predictions made by the model to the total number of predictions. Mathematically, it can be represented as:

\[Accuracy = \frac{TP+TN}{TP+FP+TN+FN}\]

\textbf{Precision}, also known as the positive predictive value, is the ratio of correctly predicted positive observations to the total predicted positive observations. High precision relates to the low false positive rate. It can be defined as:

\[Precision = \frac{TP}{TP+FP}\]

\textbf{Recall}, also known as sensitivity, hit rate, or true positive rate, is the ratio of correctly predicted positive observations to the all observations in actual class. It can be defined as:

\[Recall = \frac{TP}{TP+FN}\]

\textbf{F1 Score} in machine learning is a metric that combines both precision and recall. It is a harmonic mean of these two metrics, which means it gives much more weight to low values. As a result, the classifier will only get a high F1 Score if both recall and precision are high.

The F1 Score is particularly useful in the case of imbalanced data sets, where the negative instances vastly outnumber the positive instances. In such scenarios, a model might predict most instances as negative, leading to high accuracy but low recall. Therefore, the F1 Score is considered a better metric than accuracy in these situations.

The F1 Score is defined as:

\[F1 = \frac{2*(Recall*Precision)}{Recall + Precision}\]

In terms of True Positives (TP), False Positives (FP), and False Negatives (FN), it can also be calculated as:

\[F1 = \frac{2*TP}{2*TP+FP+FN}\]

The F1 Score ranges from 0 to 1, where 1 indicates perfect precision and recall, and 0 indicates that either the precision or the recall is zero.

In the context of multi-class classification problems, performance measures must be calculated for each class and then combined to provide an overall measure of model performance. Micro-average and macro-average are prominent methods used to accomplish this.


\textbf{Micro Average} involves aggregating the sums of False Positives (FP), False Negatives (FN), and True Positives (TP) across all classes, and then calculating Precision, Recall and F1-score. This method assigns equal weight to each instance and is thus dominated by the larger classes in imbalanced datasets. It provides a global measure used to evaluate the overall model performance.

Mathematically, micro-averaged Precision, Recall, and F1-score can be computed as follows:

\[Micro\ Precision = \frac{\sum{TP}}{\sum{TP}+\sum{FP}}\]
\[Micro\ Recall = \frac{\sum{TP}}{\sum{TP}+\sum{FN}}\]
\[Micro\ F1 = \frac{2*(Micro\ Precision * Micro\ Recall)}{Micro\ Precision + Micro\ Recall}\]

\textbf{Macro Average} calculates Precision, Recall, and F1-score for each class individually and then takes the average. This method assigns equal weight to each class, so its result is not dominated by any class, even in imbalanced datasets. It provides a local measure used to evaluate the average performance of the model across different classes.

Mathematically, macro-averaged Precision, Recall, and F1-score can be computed as follows:

\[Macro\ Precision = \frac{\sum{Precision\text{ of each class}}}{\text{Number of classes}}\]
\[Macro\ Recall = \frac{\sum{Recall\text{ of each class}}}{\text{Number of classes}}\]
\[Macro\ F1 = \frac{\sum{F1\text{ of each class}}}{\text{Number of classes}}\]




\section{Experiments and Analysis}

\subsection{Experiments and Results}

In this study, we utilized two independent datasets, OPP-115 and PPGDPR. The OPP-115 is categorized into ten classes on a paragraph basis, as shown in Table \ref{table-categories-opp-115}. The PPGDPR, on the other hand, is divided into ten classes on a sentence basis, as illustrated in Table \ref{table-categories-ppgdpr}.

\textbf{Special Process of "Other"} Similar to Polisis\cite{harkous2018polisis} and PPGDPR\cite{liu2021have}, we treated the "Other" category specially in the work, as this label's text primarily refers to introductory statements and categories not covered. For OPP-115\cite{wilson2016creation}, when handling classification results, we directly ignored instances where the human annotator labeled the text as "Other", but the Large Language Model (LLM) believed it fell into one of the other nine categories. We retained instances where LLM also identified the text as "Other". For PPGDPR\cite{liu2021have}, Out of a total of 36,610 sentences, the "Other" category accounts for 30,699. Therefore, mirroring the actions taken by Liu et al., we directly discard the "Other" category. This implies that the overall data volume is approximately around 5,000.

\textbf{Zero-shot or Few-shot?} In our research process, we implemented an A/B testing methodology to experiment with various prompts, with a specific focus on examining the impact of a select number of few-shot prompts. A/B testing, a common practice in machine learning and AI development, involves comparing two versions of a component to determine which performs better. In this context, A/B testing allowed us to directly compare the effectiveness of different prompt schemes. Our A/B tests primarily revolved around few-shot prompts. Few-shot learning is a concept in machine learning where the aim is to design machine learning algorithms that can learn useful information from a small number of examples - hence the term "few-shot". Thus, few-shot prompts are designed to train the model quickly with minimal input. However, the results from the A/B tests showed that the use of few-shot prompts did not provide a significant improvement in the accuracy of the model's classifications. On the contrary, these prompts led to substantial consumption of tokens, which can lead to inefficiencies in the model's processing abilities and resource usage. Taking these factors into consideration, we ultimately decided to adopt a zero-shot prompt scheme. In contrast to few-shot learning, zero-shot learning involves training a model to accurately classify data it has not been explicitly trained on. Considering the huge amount of data behind the LLM, zero-shot can also achieve good results, as shown in the following results.

\textbf{Measure of Accuracy} The OPP-115 dataset itself, as well as subsequent studies, do not provide a detailed explanation of the definition of accuracy. We posit that for datasets with multiple annotations for single segment or sentence, such as OPP-115\cite{wilson2016creation}, APP-350\cite{zimmeck2019maps}, etc., machine learning models, neural networks, or large language models usually yield the most fitting category. Due to the multiple perspectives involved in this process, the resulting classification tags assigned to each segment could potentially be varied, reflecting the different interpretations and focus of each annotator. When it came to evaluating the performance of the LLM in terms of output accuracy, we adopted a flexible criterion. Specifically, if the label generated by the ML/NN/LLM for a given segment was found to be among the set of labels assigned by the human experts, we deemed the classification as successful. This approach allowed for a measure of interpretive flexibility, recognizing that in complex texts such as privacy policies, multiple valid interpretations and consequently classification labels, can coexist. For datasets characterized by a single annotation, like the PPGDPR as described by Liu et al.\cite{liu2021have}, the criterion for determining the correctness of a classification is strictly based on the concurrence between the category predicted by the model and the one annotated by human evaluators.In this case, the model's prediction is juxtaposed with the human annotator's categorization to assess its accuracy. This is premised on the assumption that human annotators provide a gold standard against which machine-generated classifications are evaluated. Consequently, a classification is adjudged correct if, and only if, the model's predicted category aligns perfectly with the category annotated by the human.

Table \ref{classification-opp-115-llm} and Table \ref{classification-ppgdpr-llm} presents the classification results of ChatGPT, GPT-4, and Claude2 on OPP-115 and PPGDPR respectively.

\subsection{Analysis and Comparison}

For the two datasets, the number of instances in each category is significantly disparate. Under such circumstances, employing the $Micro\ Average$ as a metric to evaluate the overall model performance appears more justified. However, in their original papers, most authors only provide data based on the $Macro\ Average$ (which can also be calculated directly from the performance of each category if not provided). Therefore, to maintain a fair comparison, this study also uses $Macro\ Average$ figures.

The classification outcomes for the large language models ChatGPT, GPT-4, and Claude2 on the OPP-115 and PPGDPR datasets are respectively detailed in Table \ref{classification-opp-115-llm} and Table \ref{classification-ppgdpr-llm}. Correspondingly, for the OPP-115 dataset, the results of classifications by Polisis\cite{harkous2018polisis} and utilizing LR, SVM, and HMM\cite{wilson2016creation} are delineated in Table \ref{classification-opp-115-other}. For the PPGDPR dataset, the classification results employing SVM, LSTM, and BERT\cite{liu2021have} are depicted in Table \ref{classification-ppgdpr-other}. To provide a clearer comparison of the model performances, Table \ref{classification-comparison} visually represents the $Macro\ F1$ performances of each model for the two datasets.

\begin{table}[!h]
	\centering
	\caption{Classification $Precision$/$Recall$/$F1$(respectively abbreviated as P/R/F) for every single category, their $Macro\ Average$, and the total $Accuracy$ of OPP-115 by ChatGPT, GPT4 and Claude2.}
	\begin{tabular}{l c c c c c c c c c}
		\toprule  
    & \multicolumn{3}{c}{ChatGPT} & \multicolumn{3}{c}{GPT4} & \multicolumn{3}{c}{Claude2} \\
		\textbf{Label} & \textbf{P} & \textbf{R} & \textbf{F} & \textbf{P} & \textbf{R} & \textbf{F} & \textbf{P} & \textbf{R} & \textbf{F} \\
		\midrule  
		1st Party Collection & 0.94 & 0.90 & 0.92 & 0.98 & 0.97 & 0.97 & 0.99 & 0.65 & 0.78 \\
		3rd Party Sharing & 0.92 & 0.90 & 0.91 & 0.97 & 0.95 & 0.96 & 0.69 & 0.98 & 0.81 \\
    User Choice/Control & 0.92 & 0.90 & 0.91 & 0.92 & 0.98 & 0.95 & 0.77 & 0.63 & 0.69 \\
    Access, Edit, Deletion & 0.89 & 0.99 & 0.94 & 0.92 & 0.99 & 0.96 & 0.87 & 0.84 & 0.85 \\
    Data Retention & 0.93 & 0.96 & 0.95 & 1.00 & 0.81 & 0.89 & 1.00 & 0.96 & 0.98 \\
    Data Security & 0.79 & 0.96 & 0.86 & 0.98 & 0.97 & 0.97 & 0.84 & 0.85 & 0.85 \\
    Policy Change & 0.96 & 0.99 & 0.98 & 1.00 & 0.99 & 1.00 & 0.94 & 0.68 & 0.79 \\
    Do Not Track & 0.91 & 1.00 & 0.95 & 1.00 & 1.00 & 1.00 & 1.00 & 0.35 & 0.52 \\
    Specific Audiences & 1.00 & 0.92 & 0.96 & 1.00 & 0.95 & 0.97 & 0.93 & 0.79 & 0.86 \\
    Other & 0.92 & 1.00 & 0.96 & 0.99 & 1.00 & 0.99 & 0.96 & 1.00 & 0.98 \\
    \hline
    $Accuracy$ & & & 0.92 & & & 0.97 & & & 0.81 \\
    $Marco\ Average$ & 0.92 & 0.95 & 0.93 & 0.98 & 0.96 & 0.97 & 0.90 & 0.77 & 0.81 \\
		\bottomrule 
	\end{tabular}
  \label{classification-opp-115-llm}
\end{table}

\begin{table}[!h]
	\centering
	\caption{Classification $Precision$/$Recall$/$F1$(respectively abbreviated as P/R/F) for every single category, their $Macro\ Average$, and the total $Accuracy$ of PPGDPR by ChatGPT, GPT4 and Claude2.}
	\begin{tabular}{l c c c c c c c c c}
		\toprule  
    & \multicolumn{3}{c}{ChatGPT} & \multicolumn{3}{c}{GPT4} & \multicolumn{3}{c}{Claude2} \\
		\textbf{Label} & \textbf{P} & \textbf{R} & \textbf{F} & \textbf{P} & \textbf{R} & \textbf{F} & \textbf{P} & \textbf{R} & \textbf{F} \\
		\midrule  
    Collect Personal Information & 0.82 & 0.89 & 0.85 & 0.83 & 0.94 & 0.88 & 0.77 & 0.35 & 0.48 \\
    Data Retention Period & 0.75 & 0.96 & 0.84 & 0.92 & 0.93 & 0.92 & 0.90 & 0.15 & 0.26 \\
    Data Processing Purposes & 0.92 & 0.79 & 0.85 & 0.94 & 0.85 & 0.89 & 0.90 & 0.23 & 0.36 \\
    Contact Details & 0.86 & 0.90 & 0.88 & 0.95 & 0.91 & 0.93 & 0.96 & 0.54 & 0.69 \\
    Right to Access & 0.39 & 0.85 & 0.54 & 0.40 & 0.92 & 0.55 & 0.11 & 0.54 & 0.19 \\
    Right to Rectify or Erase & 0.86 & 0.72 & 0.78 & 0.89 & 0.75 & 0.81 & 0.83 & 0.41 & 0.55 \\
    Right to Restrict of Processing & 0.88 & 0.77 & 0.82 & 0.81 & 0.88 & 0.84 & 0.48 & 0.62 & 0.54 \\
    Right to Object to Processing & 0.85 & 0.84 & 0.84 & 0.85 & 0.83 & 0.84 & 0.09 & 0.89 & 0.16 \\
    Right to Data Portability & 0.96 & 0.65 & 0.77 & 0.96 & 0.62 & 0.76 & 0.30 & 0.57 & 0.40 \\
    Right to Lodge a Complaint & 0.96 & 0.94 & 0.95 & 0.96 & 0.95 & 0.95 & 0.34 & 0.97 & 0.51 \\
    \hline
    $Accuracy$ & & & 0.84 & & & 0.87 & & & 0.38 \\
    $Marco\ Average$ & 0.82 & 0.83 & 0.81 & 0.85 & 0.86 & 0.84 & 0.57 & 0.53 & 0.41 \\
		\bottomrule 
	\end{tabular}
  \label{classification-ppgdpr-llm}
\end{table}

\begin{table}[!h]
	\centering
	\caption{Classification $Precision$/$Recall$/$F1$(respectively abbreviated as P/R/F) for every single category, and their $Macro\ Average$ of OPP-115 by Polisis\cite{harkous2018polisis}, LR, SVM and HMM\cite{wilson2016creation}.}
	\begin{tabular}{l p{0.6cm}<{\centering} p{0.6cm}<{\centering} p{0.6cm}<{\centering} p{0.6cm}<{\centering} p{0.6cm}<{\centering} p{0.6cm}<{\centering} p{0.6cm}<{\centering} p{0.6cm}<{\centering} p{0.6cm}<{\centering} p{0.6cm}<{\centering} p{0.6cm}<{\centering} p{0.6cm}<{\centering}}
		\toprule  
    & \multicolumn{3}{c}{Polisis} & \multicolumn{3}{c}{LR} & \multicolumn{3}{c}{SVM} & \multicolumn{3}{c}{HMM} \\

		\textbf{Label} & \textbf{P} & \textbf{R} & \textbf{F} & \textbf{P} & \textbf{R} & \textbf{F} & \textbf{P} & \textbf{R} & \textbf{F} & \textbf{P} & \textbf{R} & \textbf{F}  \\
		\midrule  
    1st Party Collection & 0.79 & 0.79 & 0.79 & 0.73 & 0.67 & 0.70 & 0.76 & 0.73 & 0.75 & 0.69 & 0.76 & 0.72 \\
    3rd Party Sharing & 0.79 & 0.80 & 0.79 & 0.64 & 0.63 & 0.63 & 0.67 & 0.73 & 0.70 & 0.63 & 0.61 & 0.62 \\
    User Choice/Control & 0.74 & 0.74 & 0.74 & 0.45 & 0.62 & 0.52 & 0.65 & 0.58 & 0.61 & 0.47 & 0.33 & 0.39 \\
    Access, Edit, Deletion & 0.89 & 0.75 & 0.80 & 0.47 & 0.71 & 0.57 & 0.67 & 0.56 & 0.61 & 0.48 & 0.42 & 0.45 \\
    Data Retention & 0.83 & 0.66 & 0.71 & 0.10 & 0.35 & 0.16 & 0.12 & 0.12 & 0.12 & 0.08 & 0.12 & 0.09 \\
    Data Security & 0.88 & 0.83 & 0.85 & 0.48 & 0.75 & 0.59 & 0.66 & 0.67 & 0.67 & 0.67 & 0.53 & 0.59 \\
    Policy Change & 0.95 & 0.84 & 0.88 & 0.59 & 0.83 & 0.69 & 0.66 & 0.88 & 0.75 & 0.52 & 0.68 & 0.59 \\
    Do Not Track & 0.94 & 0.97 & 0.95 & 0.45 & 1.0 & 0.62 & 1.0 & 1.0 & 1.0 & 0.45 & 0.40 & 0.41 \\
    Specific Audiences & 0.96 & 0.94 & 0.95 & 0.49 & 0.69 & 0.57 & 0.70 & 0.70 & 0.70 & 0.67 & 0.66 & 0.66 \\
    \hline
    $Marco\ Average$ & 0.85 & 0.79 & 0.81 & 0.49 & 0.69 & 0.56 & 0.65 & 0.66 & 0.66 & 0.52 & 0.50 & 0.50\\
		\bottomrule 
	\end{tabular}
  \label{classification-opp-115-other}
\end{table}

\begin{table}[!h]
	\centering
	\caption{Classification $Precision$/$Recall$/$F1$(respectively abbreviated as P/R/F) for every single category, and their $Macro\ Average$ of PPGDPR by SVM, LSTM and BERT\cite{liu2021have}.}
	\begin{tabular}{l c c c c c c c c c}
		\toprule  
    & \multicolumn{3}{c}{SVM} & \multicolumn{3}{c}{LSTM} & \multicolumn{3}{c}{BERT} \\
		\textbf{Label} & \textbf{P} & \textbf{R} & \textbf{F} & \textbf{P} & \textbf{R} & \textbf{F} & \textbf{P} & \textbf{R} & \textbf{F} \\
		\midrule  
    Collect Personal Information & 0.76 & 0.05 & 0.10 & 0.49 & 0.49 & 0.49 & 0.56 & 0.56 & 0.57 \\
    Data Retention Period & 0.84 & 0.33 & 0.47 & 0.62 & 0.49 & 0.55 & 0.69 & 0.73 & 0.71 \\
    Data Processing Purposes & 0.82 & 0.03 & 0.06 & 0.61 & 0.46 & 0.52 & 0.65 & 0.57 & 0.60 \\
    Contact Details & 0.86 & 0.47 & 0.60 & 0.76 & 0.69 & 0.72 & 0.85 & 0.73 & 0.79 \\
    Right to Access & 0.71 & 0.36 & 0.47 & 0.66 & 0.50 & 0.57 & 0.65 & 0.61 & 0.63 \\
    Right to Rectify or Erase & 0.82 & 0.40 & 0.54 & 0.72 & 0.67 & 0.69 & 0.70 & 0.70 & 0.70 \\
    Right to Restrict of Processing & 0.84 & 0.50 & 0.63 & 0.78 & 0.60 & 0.68 & 0.84 & 0.76 & 0.80 \\
    Right to Object to Processing & 0.89 & 0.46 & 0.61 & 0.76 & 0.64 & 0.69 & 0.78 & 0.64 & 0.71 \\
    Right to Data Portability & 0.84 & 0.69 & 0.76 & 0.75 & 0.71 & 0.73 & 0.82 & 0.83 & 0.82 \\
    Right to Lodge a Complaint & 0.91 & 0.72 & 0.81 & 0.81 & 0.75 & 0.78 & 0.83 & 0.86 & 0.84 \\
    \hline
    $Marco\ Average$ & 0.83 & 0.40 & 0.50 & 0.70 & 0.60 & 0.64 & 0.73 & 0.70 & 0.72 \\
		\bottomrule 
	\end{tabular}
  \label{classification-ppgdpr-other}
\end{table}

\begin{table}[!h]
  \centering
  \caption{Performance($Macro\ F1$) comparison between different models.}
  \begin{tabular}{l p{1.6cm}<{\centering} p{1cm}<{\centering} p{1.5cm}<{\centering} p{1cm}<{\centering} p{0.8cm}<{\centering} p{1cm}<{\centering} p{1cm}<{\centering} p{1cm}<{\centering} p{1cm}<{\centering}}
    \toprule
    & \textbf{ChatGPT} & \textbf{GPT4} & \textbf{Claude2} & \textbf{Polisis} & \textbf{LR} & \textbf{SVM} & \textbf{HMM} & \textbf{LSTM} & \textbf{BERT} \\
    \midrule
    \textbf{OPP-115} & 0.93 & \textbf{0.97} & 0.81 & 0.81 & 0.56 & 0.66 & 0.50 & - & - \\
    \textbf{PPGDPR} & 0.81 & \textbf{0.84} & 0.41 & - & - & 0.50 & - & 0.64 & 0.72 \\
    \bottomrule
  \end{tabular} \\
  \label{classification-comparison}
\end{table}

The results indicate a superior performance by large language models, particularly GPT-4, on the privacy policy classification tasks OPP-115 and PPGDPR. This analysis will focus on the advantages these models seem to hold over other machine learning and AI techniques.

Firstly, the $Precision$, $Recall$, and $F1$ scores of GPT-4 and ChatGPT on both tasks are significantly higher than those of Polisis, LR, SVM, HMM, LSTM, and BERT. With respect to the Claude2 model, it demonstrates a performance nearly equivalent to that of traditional machine learning models on the OPP-115 dataset, but exhibits a marginally inferior performance on the PPGDPR task.

GPT-4's superior performance can be attributed to the transformer-based architecture it uses, which allows for a better understanding and generation of natural language. It can handle long-range dependencies in text and learn semantic patterns across large datasets. This is particularly beneficial in tasks like privacy policy classification where understanding context and semantics is crucial. Moreover, GPT-4 and ChatGPT's training on vast amounts of data allows them to learn a wide array of patterns and nuances in human language. This enables them to better generalize and predict in unseen situations compared to models trained on smaller datasets, such as Claude2.

The lower performance of traditional machine learning models like LR, SVM, and HMM could be due to their inability to handle highly dimensional and sequential data as effectively as deep learning models. They might struggle to capture the intricate patterns and dependencies present in natural language. In contrast, LSTM and BERT, while being neural network models, still underperform compared to GPT-4. LSTM's limitation might lie in its inability to handle extremely long sequences due to vanishing gradient problems. On the other hand, while BERT also uses transformers, it is a smaller model compared to GPT-4 and might lack the same depth and breadth of training data.

In conclusion, the robust architecture and extensive training data of large language models like GPT-4 make them highly effective in complex natural language processing tasks such as privacy policy classification. Further research and development in this area can potentially result in even more powerful models.

\section{Discussion and Conclusion}

In our work, we delve into the exploration of the potential of large language models, specifically focusing on their application in the classification of privacy policies. This avenue of research is of critical importance in the contemporary digital landscape, where privacy policies play a pivotal role yet are often complex and difficult to interpret by the layperson.

To our knowledge, this investigation is pioneering in its focus as it presents the first detailed exploration into the capabilities of large language models, such as ChatGPT and GPT-4, within the specific context of privacy policy analysis and categorization tasks. These models, with their advanced understanding and processing of natural language, present an intriguing potential for enhancing our ability to parse and classify these often verbose and convoluted policy documents.

Our experimental results offer compelling evidence of the proficiency of \textbf{PolicyGPT}. Both ChatGPT and GPT-4 exhibited remarkable performance in the analysis and categorization of privacy policies, significantly surpassing that of the baseline machine learning and neural network models. This success corroborates the potential of leveraging the advanced language understanding capabilities inherent in these models for sophisticated text processing tasks. This ability not only facilitates a more streamlined and efficient analysis process, but also generates valuable data that can serve as input for subsequent and more detailed analyses.

Looking towards the future, the potential for integrating the classification capabilities of these large language models with other analysis models appears substantial. Such synergistic combinations could unlock new approaches to privacy policy analysis, leading to more accurate, efficient, and nuanced understanding of these critical documents. This, in turn, could aid in enhancing transparency and accountability in the digital privacy landscape.

\bibliography{LLM_refs}
\bibliographystyle{unsrt}

\section*{Appendix}

\appendix

\section{Categories and Descriptions}

\setcounter{table}{0}
\renewcommand\thetable{\Alph{section}.\arabic{table}}
\begin{table}[!h]
	\centering
	\caption{Categories in OPP-115.}
  \label{table-categories-opp-115}
	\begin{tabular}{l p{9cm}}
		\toprule  
		\textbf{Category} & \textbf{Description} \\
		\midrule  
		First Party Collection/Use & how and why a service provider collects user information. \\
    Third Party Sharing/Collection & how user information may be shared with or collected by third parties. \\
    User Choice/Control & choices and control options available to users. \\
    User Access, Edit and Deletion & if and how users may access, edit, or delete their information. \\
    Data Retention & how long user information is stored. \\
    Data Security & how user information is protected. \\
    Policy Change & if and how users will be informed about changes to the privacy policy. \\
    Do Not Track & if and how Do Not Track signals for online tracking and advertising are honored. \\
    International and Specific Audiences & practices that pertain only to a specific group of users(e.g., children, Europeans, or California residents). \\
    Other & additional sub-labels for introductory or general text, contact information, and practices not covered by the other categories. \\
		\bottomrule 
	\end{tabular}
\end{table}

\begin{table}[!h]
	\centering
	\caption{Categories in PPGDPR.}
  \label{table-categories-ppgdpr}
	\begin{tabular}{l p{9cm}}
		\toprule  
		\textbf{Category} & \textbf{Description} \\
		\midrule  
		Collect Personal Information & Collect data subjects’ information which can identify their personal IDs. \\
    Data Retention Period & Retention period of personal information. \\
    Data Processing Purposes & The purposes of processing personal data. \\
    Contact Details & The contact details of the controller or the Data Protection Officer. \\
    Right to Access & The right (of the data subject) to request from the controller to access their personal information. \\
    Right to Rectify or Erase & The right (of the data subject) to request from the controller to rectify or erase of their personal information. \\
    Right to Restrict of Processing & The right (of the data subject) to request from the controller to restrict processing concerning the data subject. \\
    Right to Object to Processing & The right (of the data subject) to request from the controller to object to processing. \\
    Right to Data Portability & The right (of the data subject) to receive and transmit his/her personal data to another controller. \\
    Right to Lodge a Complaint & The right (of the data subject) to lodge a complaint with a supervisory authority. \\
		\bottomrule 
	\end{tabular}
\end{table}

\section{Full Prompt Text}
\label{sec:full-prompt}

\subsection{OPP-115}

I will give you the annotation scheme consists of ten data practice categories with its explanation. The annotations are to the website's privacy policy.
1. First Party Collection/Use: how and why a service provider collects user information.

2. Third Party Sharing/Collection: how user information may be shared with or collected by third parties.

3. User Choice/Control: choices and control options available to users.

4. User Access, Edit, \& Deletion: if and how users may access, edit, or delete their information.

5. Data Retention: how long user information is stored.

6. Data Security: how user information is protected.

7. Policy Change: if and how users will be informed about changes to the privacy policy.

8. Do Not Track: if and how Do Not Track signals for online tracking and advertising are honored.

9. International \& Specific Audiences: practices that pertain only to a specific group of users(e.g., children, Europeans, or California residents).

10. Other: additional sub-labels for introductory or general text, contact information, and practices not covered by the other categories.

For the text of privacy policy I show you below, please choose the most matching one from these ten categories and tell me. If you don't think there is any matching item, please output 'Other'.

\subsection{PPGDPR}

I will give you the annotation scheme consists of ten data practice categories with its explanation. The annotations are to the app's privacy policy. 

1.Collect Personal Information: Collect data subjects’ information which can identify their personal IDs. 

2.Data Retention Period: Retention period of personal information. 

3.Data Processing Purposes: The purposes of processing personal data.

4.Contact Details: The contact details of the controller or the Data Protection Officer.

5.Right to Access: The right (of the data subject) to request from the controller to access their personal information.

6.Right to Rectify or Erase: The right (of the data subject) to request from the controller to rectify or erase of their personal information.

7.Right to Restrict of Processing: The right (of the data subject) to request from the controller to restrict processing concerning the data subject.

8.Right to Object to Processing: The right (of the data subject) to request from the controller to object to processing.

9.Right to Data Portability: The right (of the data subject) to receive and transmit his/her personal data to another controller.

10.Right to Lodge a Complaint: The right (of the data subject) to lodge a complaint with a supervisory authority.

For the text of privacy policy I show you below, please choose the most matching one from these ten categories and tell me without any other description.

\end{document}